\documentclass[sigconf]{acmart}
\acmSubmissionID{412}
\setcitestyle{square}

\usepackage{amsbsy}
\usepackage[mathscr]{euscript}
\usepackage{stackrel}
\usepackage{nicefrac}
\usepackage{kotex}

\usepackage{algpseudocode}
\usepackage[ruled]{algorithm2e} 

\SetAlFnt{\small}
\SetAlCapFnt{\small}
\SetAlCapNameFnt{\small}
\SetAlCapHSkip{0pt}
\IncMargin{-\parindent}

\usepackage{dblfloatfix}
\usepackage{float}

\usepackage{subcaption}

\copyrightyear{2023}
\acmYear{2023}
\setcopyright{rightsretained}\acmConference[SA Conference Papers '23]{SIGGRAPH Asia 2023 Conference Papers}{December 12--15, 2023}{Sydney, NSW, Australia}
\acmBooktitle{SIGGRAPH Asia 2023 Conference Papers (SA Conference Papers '23), December 12--15, 2023, Sydney, NSW, Australia}
\acmPrice{15.00}
\acmDOI{10.1145/3610548.3618172}
\acmISBN{979-8-4007-0315-7/23/12}

\begin{document}

\title{Neural Spectro-polarimetric Fields}



\author{Youngchan Kim}
\email{kyc618@postech.ac.kr}
\affiliation{%
  \institution{POSTECH}
  \country{South Korea}
}

\author{Wonjoon Jin}
\email{jinwj1996@postech.ac.kr}
\affiliation{%
  \institution{POSTECH}
  \country{South Korea}
}

\author{Sunghyun Cho}
\email{s.cho@postech.ac.kr}
\affiliation{%
  \institution{POSTECH}
  \country{South Korea}
}

\author{Seung-Hwan Baek}
\email{shwbaek@postech.ac.kr}
\affiliation{%
  \institution{POSTECH}
  \country{South Korea}
}

\authorsaddresses{}

\begin{abstract}
Modeling the spatial radiance distribution of light rays in a scene has been extensively explored for applications, including view synthesis. Spectrum and polarization, the wave properties of light, are often neglected due to their integration into three RGB spectral bands and their non-perceptibility to human vision. However, these properties are known to encompass substantial material and geometric information about a scene. 
Here, we propose to model spectro-polarimetric fields, the spatial Stokes-vector distribution of any light ray at an arbitrary wavelength. We present Neural Spectro-polarimetric Fields (NeSpoF), a neural representation that models the physically-valid Stokes vector at given continuous variables of position, direction, and wavelength. NeSpoF manages inherently noisy raw measurements, showcases memory efficiency, and preserves physically vital signals — factors that are crucial for representing the high-dimensional signal of a spectro-polarimetric field.
To validate NeSpoF, we introduce the first multi-view hyperspectral-polarimetric image dataset, comprised of both synthetic and real-world scenes. These were captured using our compact hyperspectral-polarimetric imaging system, which has been calibrated for robustness against system imperfections. We demonstrate the capabilities of NeSpoF on diverse scenes. 
\end{abstract}

\begin{teaserfigure}
  \includegraphics[width=\linewidth]{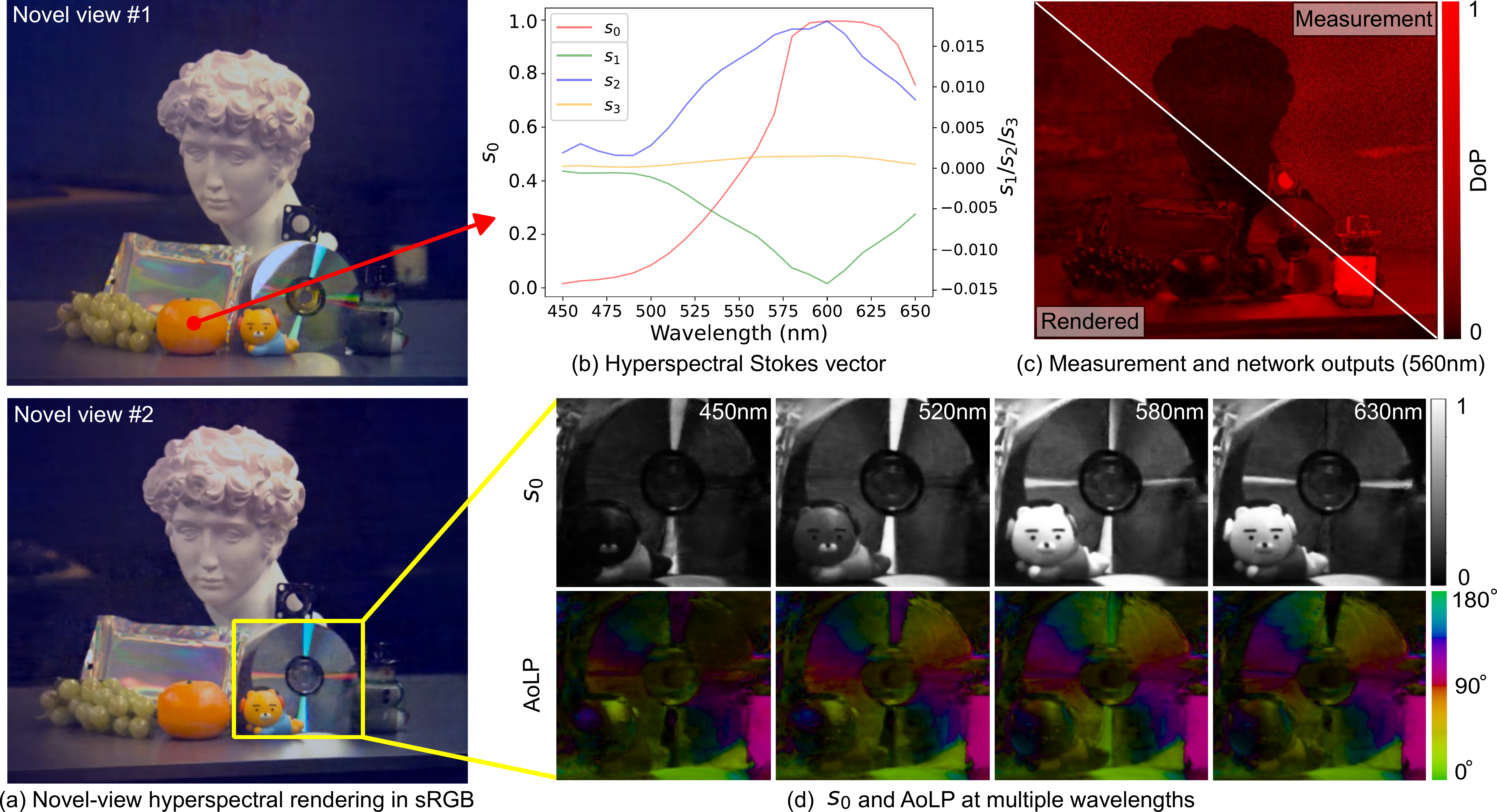}
  \centering
  \caption{ 
We present neural spectro-polarimetric fields that allow for the rendering of a scene from novel views, not limited to the synthesis of (a) RGB radiance images. We can render (b) hyperspectral Stokes vectors for each individual light ray, thus providing a comprehensive description of its spectro-polarization property. Also, (c) it can inherently reduce noise in low-signal-to-noise ratio spectro-polarimetric measurements, as shown in the degree of polarization (DoP). (d) We further show the analysis of spectro-polarimetric information for the CD that is otherwise invisible to the human eye.
}
  \label{fig:teaser}
\end{teaserfigure}

\maketitle


\section{Introduction}
\label{sec:intro}
Investigating the spatial trichromatic intensity distribution of light rays (position and direction) in a scene as a simplified plenoptic function~\cite{bergen1991plenoptic} has been extensively explored in various applications such as rendering~\cite{mildenhall2020nerf}, display~\cite{wetzstein2012tensor}, analysis~\cite{mildenhall2019local}, and imaging~\cite{levoy2006light, kalantari2016learning}. Spectrum and polarization, which are the wave properties of light, offer complementary information to the spatial distribution by disclosing the material and geometric properties of a scene. Consequently, these properties have been employed in applications such as skin analysis~\cite{zhao2009spectropolarimetric}, planet imaging in astronomy~\cite{boccaletti2012spices}, and mine detection~\cite{cheng1995aotf}.

With the objective of harnessing multiple dimensions of light concurrently, recent methods have utilized both polarization and spatial dimensions. These have found applications such as 3D reconstruction with a polarized light stage~\cite{ghosh2011multiview} and multi-view capture with a hand-held polarization camera~\cite{hwang2022sparse,dave2022pandora}. Image segmentation leveraging polarization, spectrum, and spatial dimensions was also demonstrated by Chen et al.~\shortcite{chen2022tensorial}.

While efforts have been directed towards applications, there remains a need for efficient and effective representations of high-dimensional light properties due to the curse of dimensionality affecting data size.
For instance, tensor representations~\cite{baek2021polarimetric,chen2022tensorial} have been used to handle partial dimensions of light properties, assigning each light property to a distinct dimension of the tensor. However, such direct modeling of spectrum, polarization, position, and direction can lead to prohibitive memory requirements. 
Moreover, these representations are prone to measurement noise, which can easily arise due to the cumulative effect of optical filters employed in spectral and polarimetric acquisition.

In this work, we propose to model spectro-polarimetric fields that describe the Stokes vector of a light ray, as a polarimetric representation~\cite{collett2005field,wilkie2010standardised}, at a position, direction, and wavelength.
As a representation for the spectro-polarimetric field, we present neural spectro-polarimetric fields (NeSpoF), the first neural representation specifically designed to represent \emph{physically-valid} spectro-polarimetric fields. In addition to the physical plausibility of Stokes vectors, NeSpoF handles low-SNR measurements and exhibits a compact memory footprint of 13.6\,MB, compared to a tensor-based model that consumes 58\,GB (refer to the Supplemental Document).
To evaluate NeSpoF on real-world scenes, we built a hyperspectral-polarimetric imaging system and its multi-view capture procedure, for which we propose a calibration method to account for the physical imperfections of the acquisition system.
Using the imaging system, we capture and release the first multi-view hyperspectral-polarimetric image datasets. 
On the dataset, we demonstrate the synthesis capability of NeSpoF, taking into account the view, spectrum, and polarization.

Specifically, our main contributions are described as follows: 
\begin{itemize}
    \item{We take a first step on capturing, modeling, and rendering spectro-polarimetric fields and demonstrate learning view-spectro-polarimetric variations of real-world objects.}
    \item{We develop a physically-valid neural representation, NeSpoF.}
    \item{We propose the spatially-varying spectro-polarimetric calibration method to compensate the physical imperfections of the experimental prototype.}
    \item{We release the first spectro-polarimetric multi-view dataset.}
\end{itemize}

\section{Related Work}
\label{sec:related}
\subsection{Representations of Light Properties}
Computational representation of multi-dimensional light properties is a critical consideration across visual computing applications. A prevalent method is the tensor-based approach, where each tensor dimension corresponds to a distinct light property. This approach spans the spatial tensor with two-plane methods and lumigraphs~\cite{levoy1996light, gortler1996lumigraph, sitzmann2019deepvoxels}, hyperspectral images~\cite{kim20123d}, and multi-dimensional tensors~\cite{baek2021polarimetric,chen2022tensorial}. However, the tensor representation for position, direction, spectrum, and polarization can impose an excessive memory burden due to the increased dimensionality. Moreover, it is vulnerable to the low SNR of typical hyperspectral-polarimetric measurements. 

Recently, coordinate-based neural representations~\cite{mildenhall2020nerf, suhail2022light, mildenhall2022nerf} and explicit primitives~\cite{kerbl20233d,zhang2022differentiable} modeling the spatial distribution of RGB light intensity have emerged.
Two related neural representations exist in the context of polarization and spatial dimension.
PANDORA~\cite{dave2022pandora} utilizes polarization for robust neural 3D reconstruction of specular surfaces via a polarimetric BRDF model~\cite{baek2018simultaneous}. However, the polarimetric BRDF model limits the representational capacity of PANDORA, and it does not directly model Stokes vectors under variable wavelengths. pCON~\cite{peters2023pcon} represents a polarization image based on singular-value-decomposition via a sinusoid-activation neural network. It does not model the directional dimension and is restricted to trichromatic channels.
In this work, we propose NeSpoF as a first neural representation for spectro-polarimetric fields.

\subsection{Applications using Light Properties}
The individual properties of light have been used in various applications. View synthesis and geometry reconstruction often exploit the position and direction of light~\cite{penner2017soft, mildenhall2020nerf, guo2022fast, mildenhall2019local, jin2005multi, liu2022planemvs}. The spectral dimension enables accurate detection and segmentation~\cite{imamoglu2018hyperspectral, wang2021hyperspectral, trajanovski2020tongue, aloupogianni2022hyperspectral} and estimating illumination spectrum~\cite{li2021multispectral, khan2017illuminant, an2015illumination}. Polarization provides information for shape reconstruction~\cite{kadambi2015polarized, lei2022shape, zou20203d, ba2020deep, baek2018simultaneous, ding2021polarimetric, fukao2021polarimetric}, appearance acquisition~\cite{deschaintre2021deep, kondo2020accurate}, reflection removal~\cite{nayar1997separation, wen2021polarization, yang2016method}, transparent-object segmentation~\cite{Kalra_2020_CVPR, mei2022glass}, seeing through scattering~\cite{zhou2021learning, baek2021polarimetric, liu2015polarimetric}, and tone mapping~\cite{del2019polarization}.
Simultaneously exploiting multiple light properties further enhances these capabilities, such as shape and appearance reconstruction with the spatial and polarimetric dimensions~\cite{baek2021polarimetric, dave2022pandora, riviere2017polarization}. Spectro-polarimetric analysis has been also used for object recognition~\cite{denes1998spectropolarimetric}, skin analysis~\cite{zhao2009spectropolarimetric}, and dehazing~\cite{xia2016image}.
In this work, we demonstrate view-spectrum-polarization synthesis with NeSpoF, revealing complex spectro-polarimetric variations of real-world objects at novel views.

\subsection{Acquisition of Light Properties}
Several methods have been proposed for simultaneously acquiring multiple dimensions of light.
The positional and directional dimensions can be acquired using hand-held multi-view capture of images~\cite{davis2012unstructured, mildenhall2020nerf, mildenhall2019local} or multi-view camera arrays~\cite{yang2002real, wilburn2004high, wilburn2005high}.
Capturing the positional, directional, and polarimetric dimensions has been practiced with light stages equipped with polarizers~\cite{ghosh2011multiview}, hand-held polarization cameras and flashlights~\cite{hwang2022sparse}, and polarization cameras mounted on goniometers~\cite{dave2022pandora}.
Light-field hyperspectral imaging systems~\cite{cui2021snapshot, manakov2013reconfigurable}, hyperspectral camera with a rotation stage~\cite{kim20123d}, and metalens-based snapshot spectral light-field imaging~\cite{hua2022ultra} allow for capturing the directional and spectral dimensions.
Hyperspectral-polarimetric imaging allows for capturing the spectral and polarimetric dimensions~\cite{denes1998spectropolarimetric, fan2020hyperspectral, fan2022four, altaqui2021mantis, Bai:21,chen2022tensorial, lv2020snapshot}. 
To test NeSpoF, we implement a scanning-based hyperspectral-polarimetric imaging system and propose calibrating the imperfections of the imaging system. Using the hyperspectral-polarimetric imaging system, we release the first real-world dataset of multi-view hyperspectral-polarimetric images that can be used for training NeSpoF and may spur further interest in multi-dimensional visual analysis.

\section{Background}
\label{sec:background}
\begin{figure}[t]
    \centering
    \includegraphics[width=\linewidth]{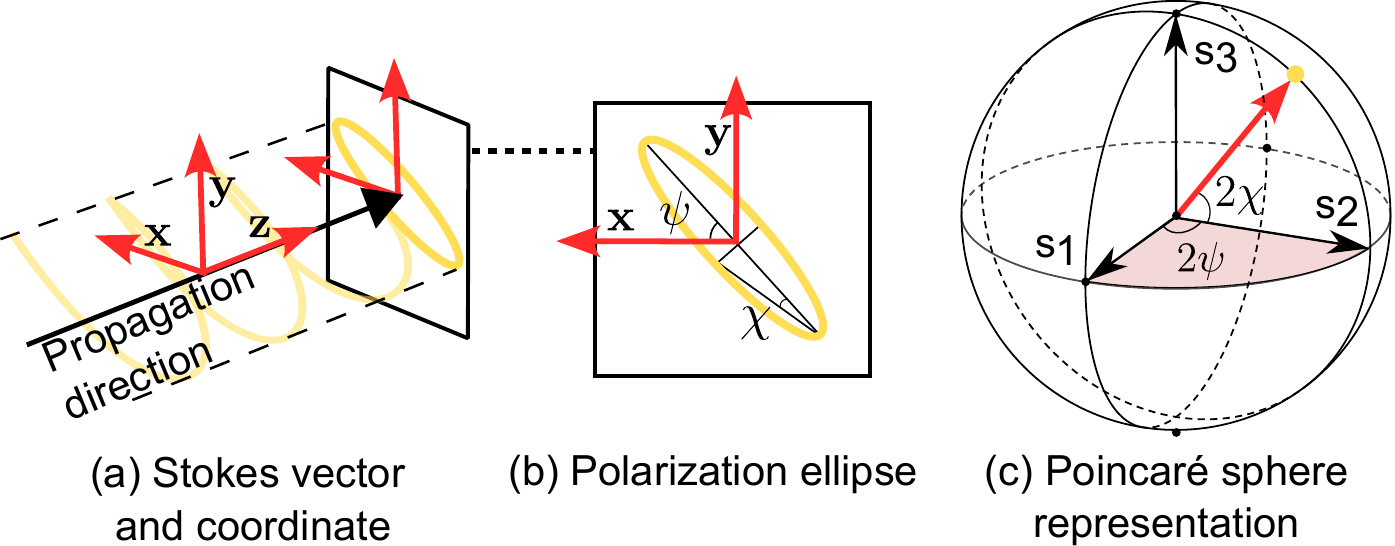}
    \caption{\label{fig:stokes_vector}
Background on polarization.
}
\vspace{-5mm}
\end{figure}
\subsection{Neural Radiance Fields}
The plenoptic function $f^p(\cdot)$ is a multi-dimensional function that describes the spatial and temporal distribution of light in a 3D space
by modeling the light intensity $L$ as a function of a 3D position $(x,y,z)$, viewing direction $(\theta, \phi)$, wavelength $\lambda$, and time $\tau$~\cite{adelson1992single}: $L = f^p(x,y,z,\theta, \phi, \lambda, \tau)$.
Assuming static scenes and constant illumination results in the static plenoptic function: $L = f^p(x,y,z,\theta, \phi, \lambda)$.

Neural radiance field (NeRF)~\cite{mildenhall2020nerf} represents the simplified plenoptic function and density of a scene as a continuous volumetric field parameterized with an MLP $F_\Theta^\text{NeRF}$ as
\begin{equation}
\label{eq:network_nerf}
{\mathbf{L}, \sigma} = F_\Theta^\text{NeRF}(x,y,z,\theta,\phi),
\end{equation}
where $\mathbf{L}=[L_R, L_G, L_B]^\intercal$ is the RGB radiance at the position $(x,y,z)$ sampled from the direction $(\theta,\phi)$.
$\sigma$ is the volume density at the position $(x, y, z)$.
$\Theta$ is the network weights.
Rendering the RGB values of a pixel at the position $\mathbf{p}$ amounts to volumetrically integrating the learned radiance $\mathbf{L}$ and density $\sigma$.

\subsection{Polarization}
\paragraph{Stokes Vector and Mueller Matrix}
Polarization is a wave property of light that describes the oscillation pattern of the electric field of light~\cite{wilkie2010standardised,collett2005field}.
The Stokes-Mueller formalism is a mathematical framework that describes the polarization state of light and how it changes.
The Stokes vector, denoted by $\mathbf{s}= [s_0, s_1, s_2, s_3]^\intercal$, describes the complete polarization state of light. The four elements of the vector represent different properties of the light.
$s_0$ represents the total radiance of the light.
$s_1$ and $s_2$ represent the differences in the radiance of linearly-polarized light at 0$^\circ$/90$^\circ$ and 45$^\circ$/-45$^\circ$ respectively.
$s_3$ represents the difference in the radiance of right- and left-circularly polarized light.
The Mueller matrix $\mathbf{M}\in \mathbb{R}^{4\times4}$ describes the change of the polarization state of a light ray by being multiplied to a Stokes vector $\mathbf{s}_\text{in}$: $\mathbf{s}_\text{out}=\mathbf{M}\mathbf{s}_\text{in}$, where $\mathbf{s}_\text{out}$ is the output Stokes vector. 

\paragraph{Coordinate of a Stokes Vector}
A Stokes vector is associated with a coordinate system $\{\mathbf{x}, \mathbf{y}, \mathbf{z}\}$, where $\mathbf{z}$ represents the direction of light propagation. 
The $\mathbf{x}$ and $\mathbf{y}$ can be chosen arbitrarily as long as they form an orthonormal coordinate system with $\mathbf{z}$, as shown in Figure~\ref{fig:stokes_vector}(a). 
This means that the polarization state of a light ray can be described differently with Stokes vectors with different elements lying at different coordinates. 

\begin{figure*}[t]
	\centering
		\includegraphics[width=0.925\linewidth]{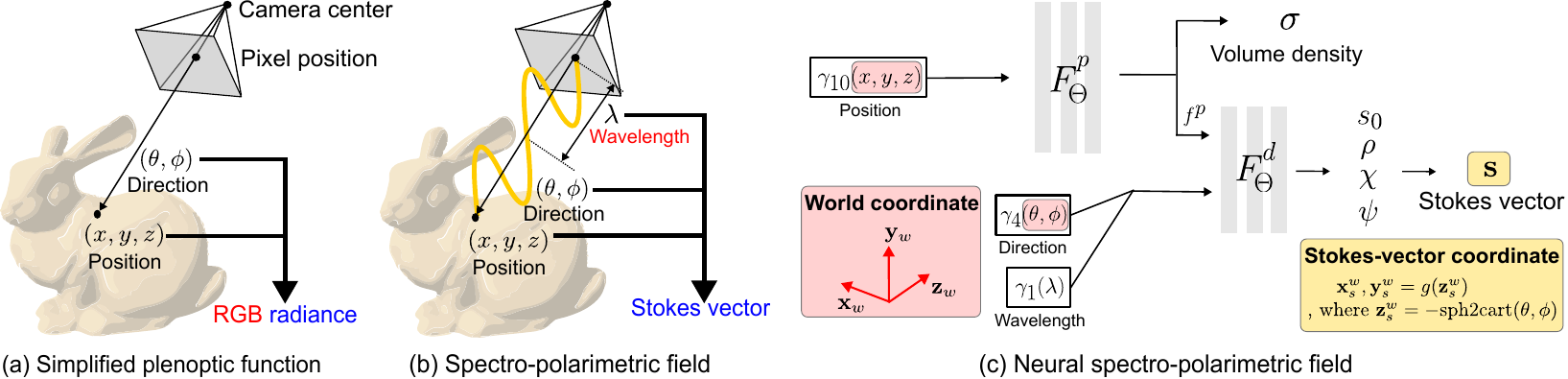}
		\caption{\label{fig:NeSpoF}
{(a) NeRF models a simplified plenoptic function to represent spatially-distributed RGB radiance. (b) The spectro-polarimetric field describes the polarization state and intensity of light as a Stokes vector per each wavelength $\lambda$, position $x, y, z$, and direction $\theta, \phi$. (c) NeSpoF describes the spectro-polarimetric field, considering the physical validity of Stokes vectors and wavelength as an input variable. See Section~\ref{sec:method} for details of NeSpoF.}
}
\end{figure*}

\paragraph{Polarization Ellipse and Poincaré Sphere}
The polarization ellipse describes the projected pattern of the oscillating electric field onto the tangent plane.
Figure~\ref{fig:stokes_vector}(b) shows the azimuth angle $\psi$, which is the angle of linear polarization, and the ellipticity $\chi$ that determines the ratio between circularity and linearity.
The Poincaré sphere, shown in Figure~\ref{fig:stokes_vector}(c), depicts polarization in three-dimensional space with the first, second, and third Stokes-vector elements normalized by the total radiance: $\{{s_1/s_0}, {s_2/s_0}, {s_3/s_0}\}$.

\paragraph{Visualization of Stokes Vectors}
For visualizing Stokes vectors, we use degree of polarization (DoP), angle of linear polarization (AoLP), type of polarization (ToP), and chirality of polarization (CoP)~\cite{wilkie2010standardised}. 
DoP describes the ratio of polarized light radiance compared to the total radiance, AoLP is the angle of the linearly-polarized component with respect to the $\mathbf{x}$ axis of the Stokes-vector basis, corresponding to $\psi$, ToP describes the ratio of linear and circular components corresponding to $\chi$, and CoP refers to the handedness of circularly polarized light corresponding to $\chi$.

\section{Neural Spectro-polarimetric Field}
\label{sec:method}
\paragraph{Spectro-polarimetric Field}
We define spectro-polarimetric field $f(\cdot)$ as a description of the Stokes-vector field at any given position, direction, and wavelength:
$\mathbf{s} = f(x,y,z,\theta, \phi, \lambda)$,
where $\mathbf{s}$ is the Stokes vector.
Unlike the simplified plenoptic function used in NeRF~\cite{mildenhall2020nerf}, the spectro-polarimetric field incorporates wavelength as an input variable and outputs the Stokes vector. See Figure~\ref{fig:NeSpoF}.

\paragraph{NeSpoF}
We propose modeling the spectro-polarimetric field $f(\cdot)$ using NeSpoF, a neural representation that describes the Stokes vector $\mathbf{s}$ and volumetric density $\sigma$ for the input variables of position, direction, and wavelength:
\begin{equation}
\mathbf{s}, \sigma = F_\Theta(x,y,z,\theta, \phi, \lambda).
\end{equation}
Using NeSpoF, we can render the Stokes vector $\mathbf{s}_\text{rendered}$ from a ray $\mathbf{r}$ emitted from a particular camera pixel at a wavelength $\lambda$ by applying volumetric rendering as follows:
\begin{align}
\label{eq:vol_render_neural_wave_extended_plenoptic_fields}
&\mathbf{s}_\text{rendered}(\mathbf{r},\lambda)=\int_{t_n}^{t_f} \underbrace{\exp \left(-\int_{t_n}^{t}{\sigma(\mathbf{r}(s))ds} \right)}_{\text{accumulated transmittance}} \underbrace{ \sigma\left(\mathbf{r}(t)\right) \mathbf{s}(\mathbf{r}(t),\mathbf{d},\lambda)}_{\text{transmitted Stokes vector}}dt,
\end{align}
where $t_n$ and $t_f$ are the near and far bounds.
For details on accumulated transmittance and integral form, we refer to the linear differential formulation of volumetric rendering~\cite{kajiya1984ray}.
The ray is modeled as $\mathbf{r}(t)=\mathbf{o}+t\mathbf{d}$ with respect to sample $t$.
$\mathbf{o}$ is the center of the camera and $\mathbf{d}$ is the direction vector of the ray.

Note that this linear differential formulation has been previously applied to the radiance of light~\cite{kajiya1984ray}.
Here, we instead apply it to Stokes vectors with an assumption that the polarization state of light does not change during transmittance.

\paragraph{Model Design}
We design NeSpoF with a positional MLP, $F_\Theta^p$, and a spectro-directional MLP, $F_\Theta^d$, as shown in Figure~\ref{fig:NeSpoF}(c).
The positional MLP adopts the original NeRF design. It estimates the volumetric density $\sigma$ at the position $(x,y,z)$ and extracts the positional Stokes-vector feature $f^p$.
The positional encoding $\gamma_k(x)$~\cite{tancik2020fourier} enables learning high-frequency details of the Stokes-vector distribution with respect to the position:
\begin{equation}
\label{eq:pos_enc}
\gamma_k(x) = [x, \sin(2^0\pi x), \cos(2^0\pi x), ..., \sin(2^k\pi x), \cos(2^k\pi x))],
\end{equation}
where $k$ is a hyperparameter. For position, $k=10$ is used.

We now turn to model a physically-valid Stokes vector at a wavelength $\lambda$ using the spectro-directional MLP.
The spectro-directional MLP $F_\Theta^d$ takes the positional feature $f^p$, directional features, and wavelength features and outputs intermediate polarimetric features.
Below, we describe the key factors of spectro-directional MLP. 

\paragraph{Wavelength as an Input}
We use wavelength $\lambda$ as an input to the spectro-directional MLP in addition to the direction variables $\theta$ and $\phi$, unlike NeRF that takes only the direction variables as input and outputs as three RGB radiance values.
In this way, we can query the Stokes vector from NeSpoF at an arbitrary wavelength $\lambda$, suitable for spectral analysis and measurements of the Stokes vectors.
We apply the positional encoding of Equation~\eqref{eq:pos_enc} to the wavelength with the hyperparameter value $k=1$. 

\paragraph{Physically-valid Stokes Vector}
It is crucial to ensure that NeSpoF outputs \emph{physically-valid} Stokes vectors.
A Stokes vector is physically valid if the following condition is met: $s_0^2 \geq s_1^2+s_2^2+s_3^2$~\cite{collett2005field}.
A na\"ive approach of modeling the output of the spectro-directional MLP as Stokes-vector elements fails to guarantee the physical validity of the output Stokes vector.
Hence, we propose to use intermediate polarimetric properties as outputs: $X_0$, $X_1$, $X_2$, $X_3$.
We then map the values to the total intensity $s_0$, DoP $\rho$, ellipticity $\chi$, and azimuth angle $\psi$ as follows:
\begin{align}
\label{eq:activation_stokes}
s_0 = S(X_0),\quad \rho = S(X_1), \quad \chi = X_2, \quad \psi = X_3,
\end{align}
where $S(\cdot)$ is the sigmoid activation function that bounds the intensity $s_0$ and the DoP $\rho$ from zero to one.
Note that we do not apply bounding functions to the azimuth angle $\psi$ and the ellipticity $\chi$, which naturally wrap within $2\pi$ by sinusoidal functions.
We then construct a physically-valid Stokes vector $\mathbf{s}$ as follows:
\begin{align}
\label{eq:poincare_sphere}
s_1 = s_0 \rho \cos 2\chi \cos 2\psi, 
s_2 = s_0 \rho \cos 2\chi \sin 2\psi, 
s_3 = s_0 \rho \sin 2\chi.
\end{align}

\paragraph{Coordinate of Output Stokes Vector}
NeSpoF models a field of Stokes vectors, which must be associated with a Stokes-vector coordinate for each light ray.
{Given the position $(x,y,z)$ and the direction $(\theta, \phi)$ of a light ray defined on the world coordinate system $\{\mathbf{x}_w, \mathbf{y}_w, \mathbf{z}_w\}$, we denote the coordinate system of the output Stokes vector as $\{\mathbf{x}_s^w, \mathbf{y}_s^w, \mathbf{z}_s^w\}$, where $\mathbf{z}_s^w$ is aligned with the reversed direction vector corresponding to $(\theta, \phi)$: $\mathbf{z}_s^w = -\mathrm{sph2cart}(\theta, \phi)$.} $\mathrm{sph2cart}(\cdot)$ is the conversion function that maps a spherical coordinate to a Cartesian coordinate.
In order to ensure that the coordinate system for the Stokes vector, $\mathbf{s}$, is well-defined, we compute the other two axes $\mathbf{x}_s^w$ and $\mathbf{y}_s^w$ deterministically to form an orthonormal coordinate system using an analytic method denoted as $g$~\cite{duff2017building}:
$\mathbf{x}_s^w, \mathbf{y}_s^w = g(\mathbf{z}_s^w)$.
This rule, also shown in Figure~\ref{fig:NeSpoF}(c), for determining the Stokes-vector coordinate enables us to use multi-view hyperspectral images of which Stokes vectors are represented in a local coordinate. 
Refer to the Supplemental Document for details.
 
\paragraph{Training with Polarization-weighted Loss}
To train NeSpoF, we cast rays $\mathbf{r}$ and apply volumetric rendering as in Equation~\eqref{eq:vol_render_neural_wave_extended_plenoptic_fields} for the Stokes vector $\mathbf{s}^\lambda$ and density $\sigma$, minimizing the difference between the rendered and measured Stokes vectors:
\begin{align}
\label{eq:opt_neural_wave_extended_plenoptic_fields}
  \underset{\Theta}{\text{minimize}} \sum_{\mathbf{r}\in \mathcal{R}}  \sum_{\lambda}  \sum_{i=0}^3    w_i \left([\mathbf{s}_\text{meas}(\mathbf{r},\lambda)]_i -  [\mathbf{s}_\text{rendered}(\mathbf{r},\lambda)]_i \right)^2,
\end{align}
where $\mathcal{R}$ is the set of rays in a training batch and $[\mathbf{s}]_i$ retrieves the $i$-th element of the Stokes vector $\mathbf{s}$.
We use the rendered Stokes vector $\mathbf{s}_\text{rendered}$ for volumetric rendering of Equation~\eqref{eq:vol_render_neural_wave_extended_plenoptic_fields}.
We propose an adaptive weight $w_i$ for the $i$-th Stokes-vector element to account for the scale differences among the Stokes-vector elements:
\begin{align}
\label{eq:weight}
  w_i = \frac{\text{std.dev.}\left([\mathbf{s}_\text{meas}(\forall)]_0\right)}{\text{std.dev.}\left([\mathbf{s}_\text{meas}(\forall)]_i\right)},
\end{align}
where $\text{std.dev.}(\cdot)$ is the standard deviation operator. 
This strategy aims to learn all the Stokes-vector elements evenly considering their relative scales, as small-scale elements may convey physically significant meanings despite their absolute scales.
We apply the weight variance regularizer for training~\cite{mildenhall2022nerf}.

\section{Spectro-polarimetric Imaging}
\label{sec:acquisition}
To test NeSpoF on real-world scenes, we construct an imaging system and propose a calibration method to address the imperfections of the imaging system. Using the imaging system, we capture and release the first real-world multi-view hyperspectral-polarimetric image dataset.

\begin{figure}[t]
	\centering
		\includegraphics[width=\columnwidth]{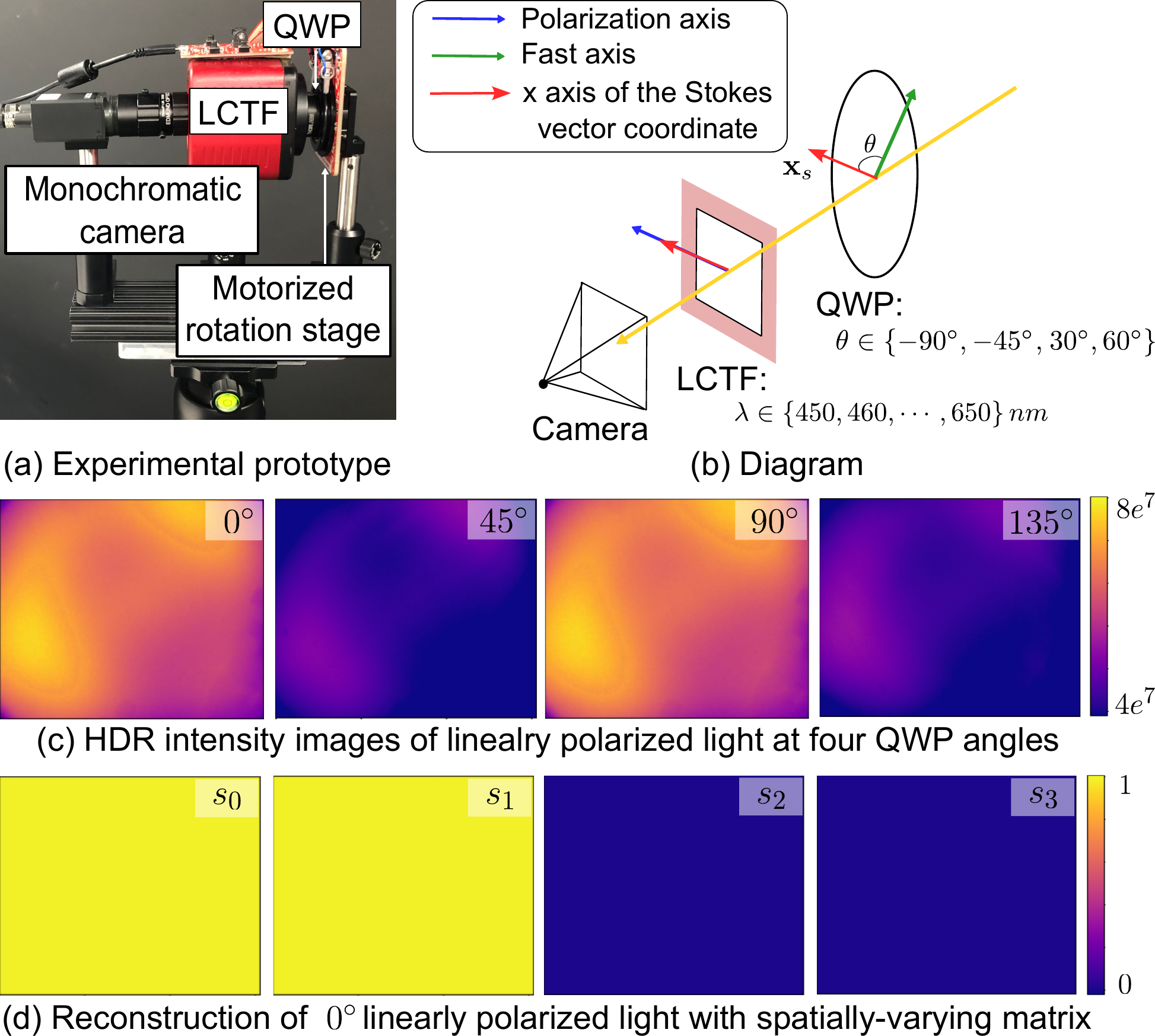}
		\caption{\label{fig:imaging system}
  {
		(a) Experimental prototype and (b) schematic diagram. We capture images with the QWP at four different angles at each wavelength. 
(c) The spatially-varying spectro-polarimetric transmission of the LCTF results in spatial non-uniformity. The images of linearly-polarized uniform light are shown at the four QWP angles. 
(d) We calibrate such non-uniformity with spatially-varying Mueller matrices, resulting in accurate polarimetric reconstruction of $0^\circ$ degree linearly-polarized light. 
}
}
\end{figure}

\subsection{Imaging}
\paragraph{Imaging Setup}
Our experimental prototype, shown in Figure~\ref{fig:imaging system}(a), consists of a monochromatic sensor equipped with a 35\,mm lens, a liquid crystal tunable filter (LCTF), and a quarter-wave plate (QWP) mounted on a motorized rotation stage. The LCTF can adjust spectral transmission. We placed a cut-off filter for the IR and UV wavelengths in front of the LCTF.

\paragraph{Image Formation}
Assume that light of a wavelength $\lambda'$ coming from a scene with a Stokes vector $\mathbf{s}({\lambda'})$ enters the QWP mounted on a rotation stage, at an angle $\theta$ as shown in Figure~\ref{fig:imaging system}(b). 
This causes the Stokes vector to change to $\mathbf{Q}(\theta,\lambda')\mathbf{s}({\lambda'})$, where $\mathbf{Q}(\theta,\lambda')$ is the Mueller matrix of the QWP when its fast axis is at an angle $\theta$.
The light then passes through the LCTF. 
At each target wavelength $\lambda \in \{450, 460, \cdots, 650\}\,nm$, we set the LCTF to maximize its transmission at the target wavelength $\lambda$ and suppress the transmission at other wavelengths. 
We model the spectral transmission and polarization change by the LCTF with the first row of the unknown LCTF Mueller matrix as $\mathbf{M}(\lambda'; \lambda)\in \mathbb{R}^{1\times4}$.
The optical cut-off filter, which has a spectral transmission of $F(\lambda')$, attenuates the radiance, and the monochromatic sensor, whose quantum efficiency is $T(\lambda')$, senses the light intensity.
In summary, we describe the image formation as:
\begin{align}
    I^{\lambda,\theta} &=  \int_{\lambda'} T(\lambda') F(\lambda')  \mathbf{M}(\lambda';\lambda) \mathbf{Q}(\theta,\lambda') \mathbf{s}({\lambda'}) d\lambda' \\
        &\approx  {T}(\lambda) {F}(\lambda) {\mathbf{M}}(\lambda;\lambda) {\mathbf{Q}}(\theta,\lambda) \mathbf{s}(\lambda),  \label{eq:image_formation_continuous}
\end{align}
where the approximation in Equation~\eqref{eq:image_formation_continuous} generally holds as the average bandwidth of the LCTF over the visible spectrum is 9.8\,nm.
We use the factory-calibrated quantum efficiency of the sensor, $T(\lambda)$ and the transmission functions of the cut-off filter, $F(\lambda)$.

\begin{figure*}[t]
	\centering
		\includegraphics[width=\linewidth]{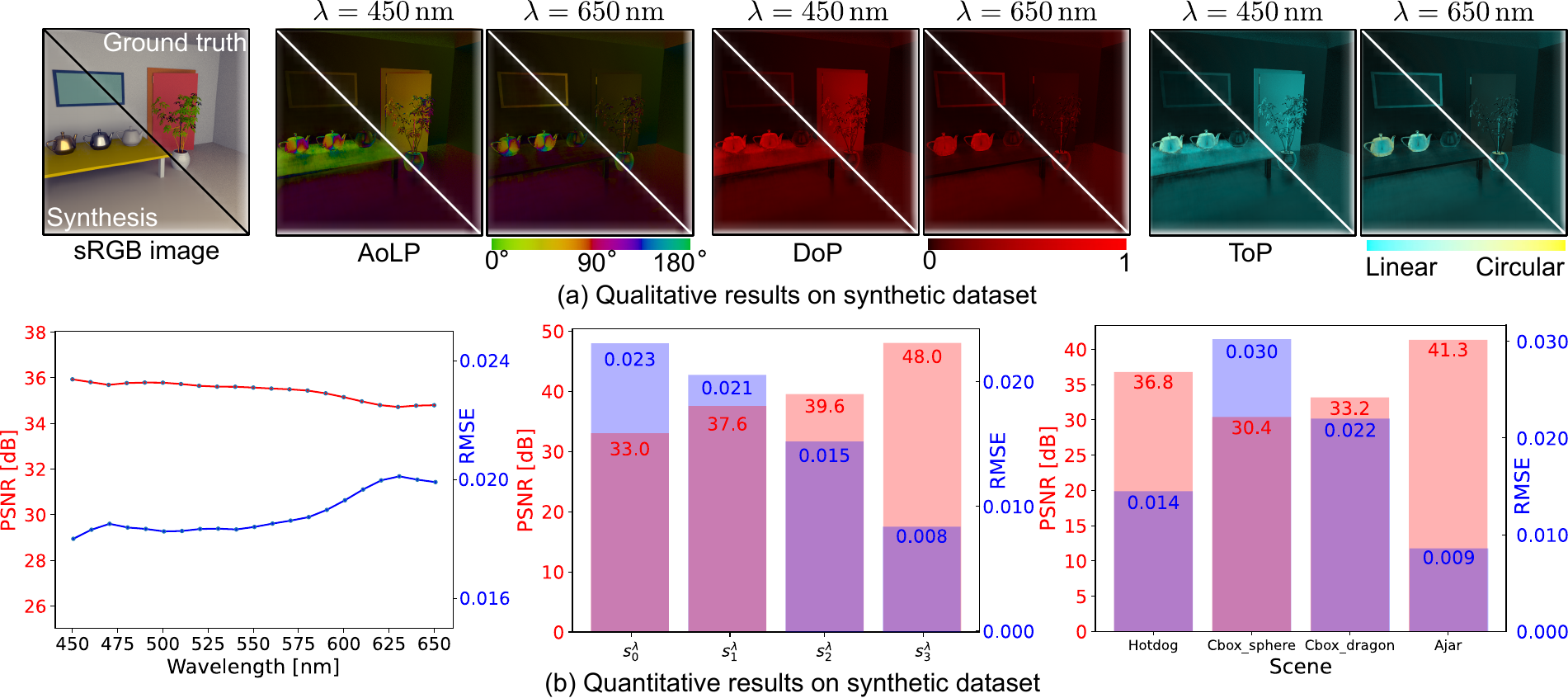}
		\caption{Synthetic evaluation. NeSpoF achieves an accurate synthesis of hyperspectral Stokes vectors at novel views, evidenced in both qualitative and quantitative analysis.
  }
		\label{fig:synthetic_results}
\end{figure*}

\begin{figure}[t]
    \centering
    \includegraphics[width=\linewidth]{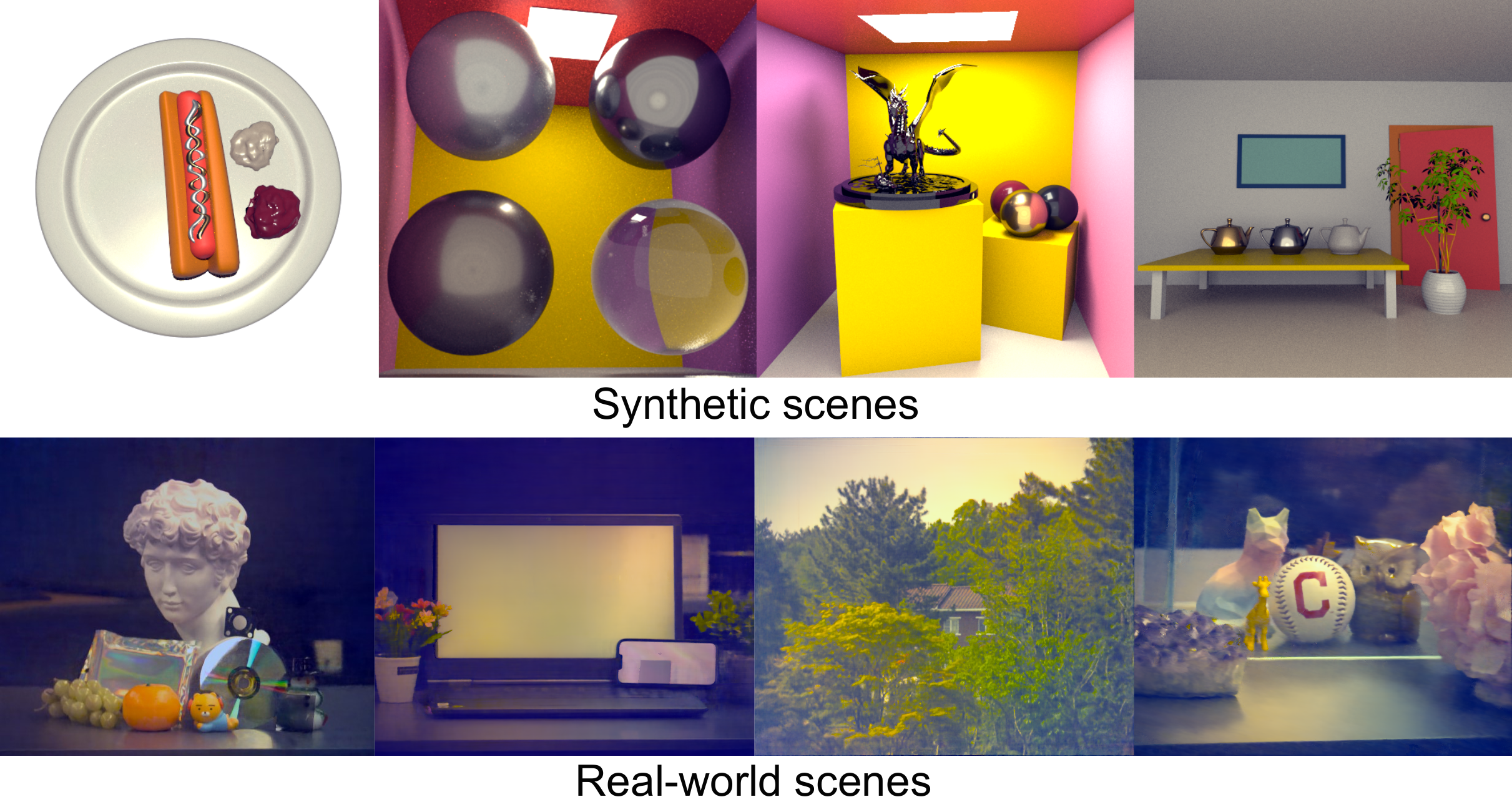}
    \caption{Multi-view hyperspectral polarimetric dataset.}
    \label{fig:dataset}
    \vspace{-5mm}
\end{figure}

\paragraph{Calibration of LCTF Modulation}
Even though we designed our imaging system using professional optical devices, the experimental prototype still presents deviation from a commonly-assumed model for the polarization modulation of LCTF, which has often been assumed to be an ideal linear polarizer~\cite{fan2023compressive}. As shown in Figure~\ref{fig:imaging system}(c), the captured intensity through our imaging system shows spatially non-uniform spectro-polarimetric variations.
To overcome this problem, we propose to optimize for spatially-varying matrices $\mathbf{M}(\lambda,p;\lambda)$ that act as polarization modulation by the LCTF, where $p$ is a pixel.
We capture 16 samples with known Stokes vectors constructed with {a} LP, a QWP, and a collimated LED light source.
We then solve for the matrix $\mathbf{M}$ as follows:
\begin{align}
\label{eq:compensation}
\mathop {\text{minimize}}\limits_{\mathbf{M}(\lambda,p)}\sum\limits_{i= 1}^{N}  \sum\limits_{k = 1}^{K} {\left( I^{\lambda,\theta_k}_{\mathrm{meas},i}(p)- T(\lambda)F(\lambda) \mathbf{M}(\lambda,p;\lambda) {\mathbf{Q}}(\theta_k,\lambda)  \mathbf{s}_{\text{u}}  \right)}^2,
\end{align}
where $I^{\lambda,\theta_k}_{\mathrm{meas},i}(p)$ is the intensity measurement of the $i$-th sample for the pixel $p$ at the rotation angle $\theta_k$ of the QWP. $\mathbf{s}_{\text{u}}$ is the Stokes vector of unpolarized light.  
We used four uniformly-sampled $\theta$ in the {$[0,\pi)$} range for this calibration.
We refer to the Supplemental Document for its normal-equation formulation. 

\paragraph{Stokes-vector Reconstruction}
Once the matrix $\mathbf{M}$ is obtained, we can reconstruct the per-pixel Stokes vector $\mathbf{s}_\text{meas}$ for a wavelength $\lambda$ by solving the least-squares problem:
\begin{align}
\label{eq:reconstruction_real}
\mathop {\text{minimize}}\limits_{\mathbf{s}_\text{meas}(\lambda)}\sum\limits_{k = 1}^K {\left( I^{\lambda,\theta_k}_{\mathrm{meas}}- c(\lambda)\mathbf{M}(\lambda;\lambda) {\mathbf{Q}}(\theta_k,\lambda) \mathbf{s}_\text{meas}(\lambda) \right)}^2,
\end{align}
where $c(\lambda)=T(\lambda)F(\lambda)$. We exclude the pixel dependency in the equation for readability.
Figure~\ref{fig:imaging system}{(d)} shows the results of the calibrated Stokes-vector reconstruction.

\begin{figure*}[t]
	\centering
		\includegraphics[width=\linewidth]{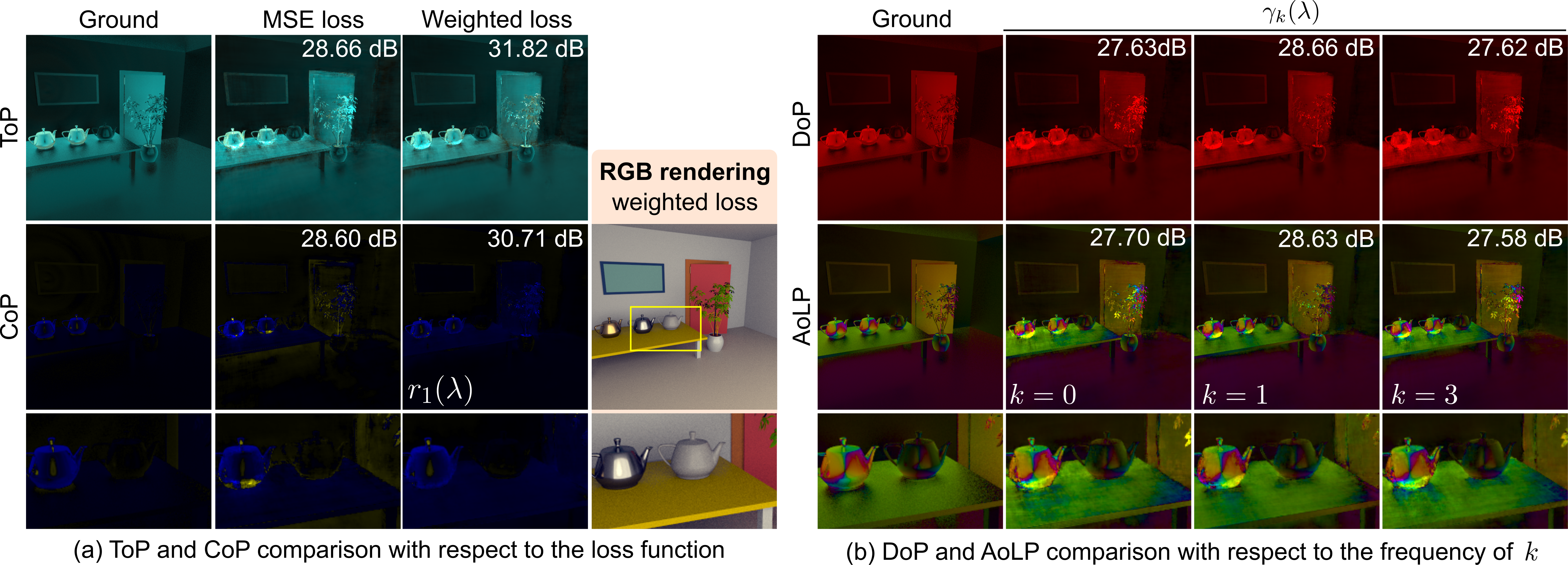}
		\caption{
    We evaluate the impacts of (a) the utilization of weighted loss and (b) the frequency $k$ of the positional encoding for wavelength. Using the weighted loss and the frequency of $k=1$ enables better polarimetric reconstruction. We report PSNRs of the estimated ToP and CoP for the former, and DoP and AoLP for the later.
    }
		\label{fig:ablation}
\end{figure*}

\begin{figure*}[t]
	\centering
		\includegraphics[width=\linewidth]{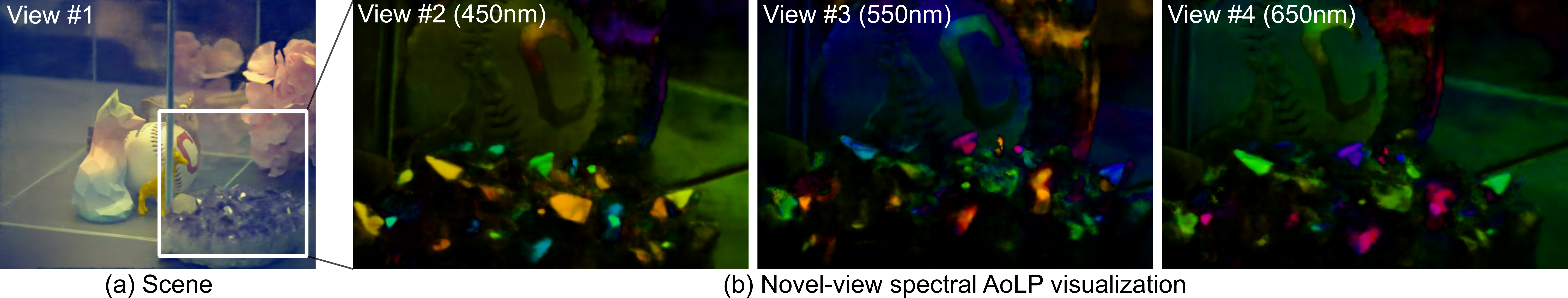}
		\caption{NeSpoF allows for synthesizing view-dependent hyperspectral-polarimetric patterns of the crystal and a baseball.  
  }
		\label{fig:real_results_crystal}
\end{figure*}

\subsection{Dataset}
Using the experimental prototype, we acquired the first multi-view hyperspectral polarimetric dataset of four real-world scenes. 
At each camera viewpoint, we capture raw measurements and reconstruct the per-pixel Stokes vector $\mathbf{s}^\lambda_\text{meas}$ using Equation~\eqref{eq:reconstruction_real}. 
It takes roughly 10\,seconds to capture a hyperspectral-polarimetric image with 21 spectral channels at four QWP angles {($-90^\circ$, $-45^\circ$, $30^\circ$, and $60^\circ$)~\cite{ambirajan1995optimum}}. 
We then perform local-to-world coordinate conversion to use it for NeSpoF.
We obtain the geometric parameters of the multiple viewpoints and the camera intrinsic parameters using a structure-from-motion method, COLMAP~\cite{schoenberger2016sfm}, for which we use the RGB intensity images obtained by applying the spectrum-to-RGB conversion as well as the gamma correction. 
To test NeSpoF on clean images without measurement noise, we also render multi-view hyperspectral polarimetric images of four synthetic scenes using Mitsuba3~\cite{jakob2022mitsuba3}.
We manually assign the multispectral polarimetric BRDFs~\cite{baek2020image} to meshes~\cite{mildenhall2020nerf,resources16}. Since only five spectral channels are available (450, 500, 550, 600, and 650\,nm) in the BRDF dataset, we fit a fourth-order polynomial function at each rendered pixel along the spectrum. 
Figure~\ref{fig:dataset} shows the thumbnails of our multi-view hyperspectral polarimetric dataset. We refer to the Supplemental Document for more details about the dataset, calibration, and coordinate conversion.

\section{Assessment}
\label{sec:assessment}
\subsection{View-spectrum-polarimetric Synthesis}
We train NeSpoF for each scene in the dataset and test the capability of NeSpoF on view-spectrum-polarization synthesis.

\paragraph{Synthetic Scenes}
We demonstrate the effectiveness of NeSpoF on the synthetic dataset.
Figure~\ref{fig:synthetic_results}(a) shows the rendered and ground-truth hyperspectral-polarimetric Stokes-vector images.
The synthesized results accurately match the ground truth such as visualized in AoLP, DoP, and ToP. 
Figure~\ref{fig:synthetic_results}(b) shows the quantitative results on all test views of the synthetic dataset. In terms of wavelength, NeSpoF obtains PSNR values of 35\,dB and an RMSE of 0.020 at 630\,nm, with better reconstructions for other wavelengths. For each Stokes-vector element, the radiance component, which usually exhibits the most complex variation, has a PSNR of 33\,dB and RMSE of 0.023, while the other elements demonstrate higher PSNRs and lower RMSEs. The scene-dependent quantitative analysis shows a minimum PSNR of 30.4\,dB.
Refer to the Supplemental Document for additional synthetic-scene results.

\begin{figure*}[t]
	\centering
		\includegraphics[width=\linewidth]{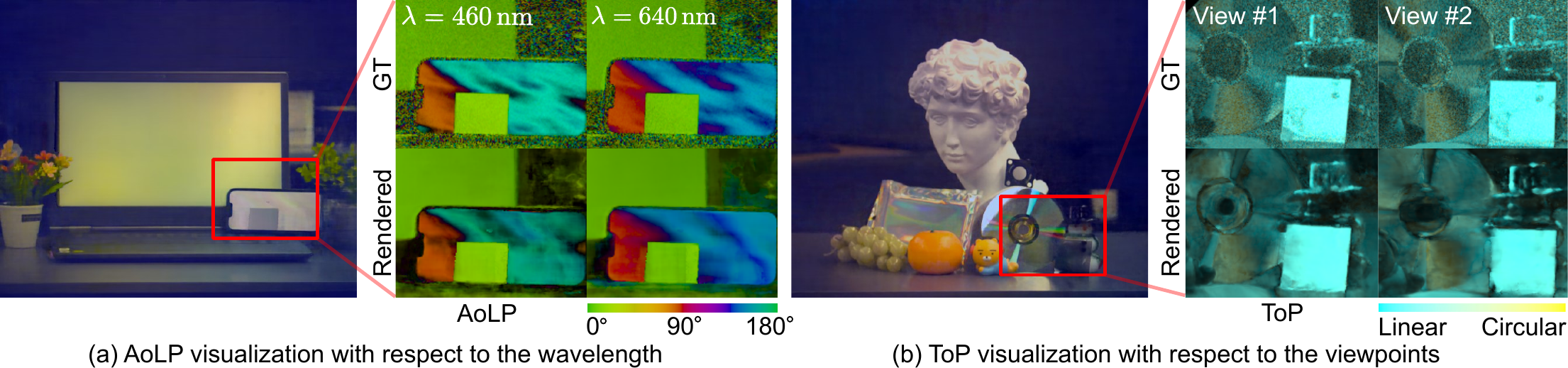}
		\caption{
We demonstrate the synthesis of view-spectrum-polarization using NeSpoF in two real-world scenes that exhibit unique view-spectro-polarimetric characteristics, specifically a mobile display covered with stressed protection film, a linear polarizer, and a CD. We visualize AoLP at two different wavelengths from a novel view, and ToP at different views.
  }
		\label{fig:real_results}
\end{figure*}

\paragraph{Real-world Scenes}
NeSpoF enables modeling complex view-spectro-polarimetric information of real-world scenes.
Figure~\ref{fig:real_results_crystal} demonstrates NeSpoF on synthesizing the intricate variations of AoLP from the reflections of the crystal and the baseball inside of a glass box. 
Figure~\ref{fig:real_results} shows two real-world scenes.
The first scene consists of a laptop and a mobile phone equipped with a protective film. In front of the mobile phone is a linear polarizer. The laptop display emits clean linearly polarized light whereas the light from the phone display interacts with the stress-affected protective film.
The second scene contains a CD demonstrating unique spectro-polarimetric characteristics attributed to diffraction caused by micro-scale surface structures, paired with a doll covered by a linear polarizer. 
NeSpoF accurately models the resulting complex spectro-polarimetric patterns.
Moreover, the regularization capability of NeSpoF can be observed in low SNR regions.

\begin{figure*}[t]
	\centering
		\includegraphics[width=\linewidth]{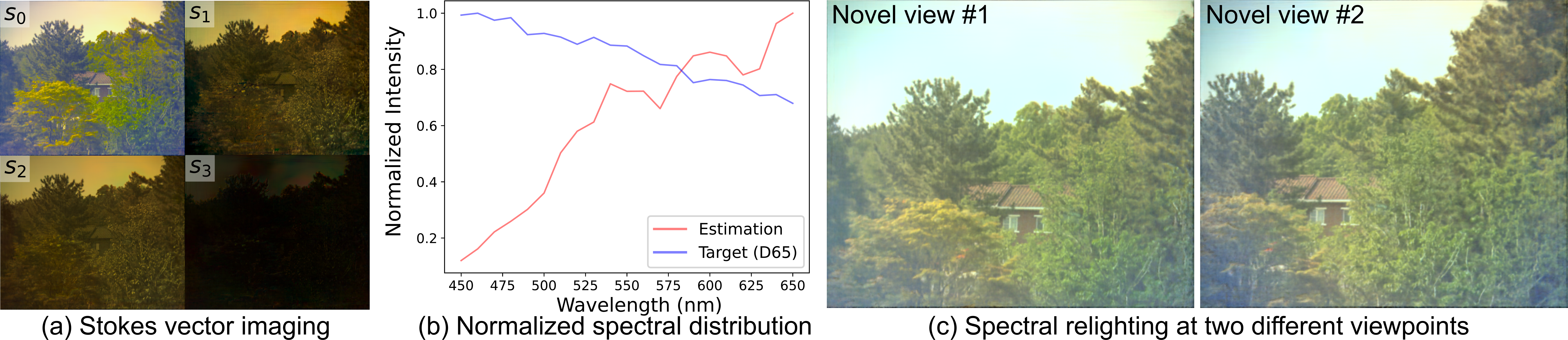}
		\caption{(a) NeSpoF facilitates the rendering of hyperspectral-polarimetric images of an outdoor scene at a novel view, revealing the partially-linear polarization state of skylight. (b) We extract the hyperspectral intensity of the skylight and (c) use it to relight the scene from two novel views under D65 illumination.
  }
		\label{fig:illum_est}
\end{figure*}

\paragraph{Novel-view Spectral Relighting}
Figure~\ref{fig:illum_est} demonstrates novel view synthesis by NeSpoF for an outdoor scene under sky lighting. This allows for analysis of the spectro-polarimetric properties of sky light, which is partially linearly polarized, as indicated by high values of $s_1$ and $s_2$. The illumination spectrum of the sky can also be extracted, as depicted in Figure~\ref{fig:illum_est}(b). This enables rendering of the hyperspectral intensity at a novel view and relighting by replacing the original illumination with the target illumination, here CIE D65 illuminant, shown in Figure~\ref{fig:illum_est}(c).

\begin{figure*}[t]
	\centering
		\includegraphics[width=\linewidth]{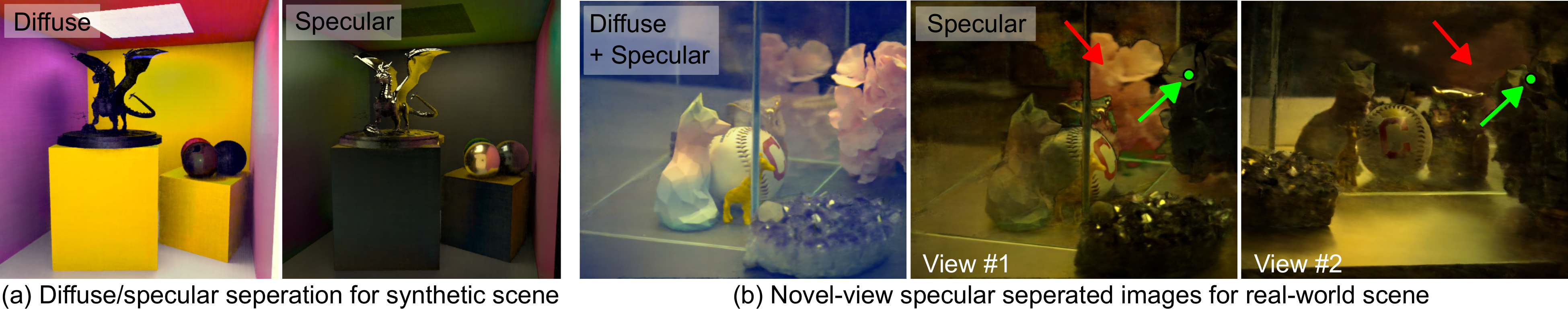}
		\caption{
{(a) NeSpoF enables the separation of unpolarized diffuse and polarized components, which include specular reflections and polarized inter-reflections for the synthetic scene. (b) For the real-world scene, we identify the different polarization states of a flower and its reflection to glass depending on reflection angles.}
 }
\label{fig:diff_spec}
\end{figure*}

\paragraph{Diffuse, Specular, Inter-reflection, and Reflection Components}
Figure~\ref{fig:diff_spec} shows the analysis of the unpolarized and polarized hyperspectral intensity with NeSpoF. This enables us to separate not only diffuse and specular components but also more complex light transport, such as the polarized inter-reflection between walls for the synthetic dragon scene.
For the real-world scene, it is observed that the diffuse component of the flower is unpolarized, and hence not visible in the specular image. However, its reflection on the glass is clearly visible in the specular image due to the strong polarization caused by Fresnel reflection near a Brewster angle. When rendering the frontal-view specular image of the glass, the strong specular signal of the flower on the glass is not visible, as the imaging geometry deviates from the Brewster-angle configuration.

\subsection{Analysis of NeSpoF}
Being the first neural model to address the modeling of hyperspectral Stokes fields, we here focus on the analysis of NeSpoF. Comparative studies can be found in the Supplemental Document.

\paragraph{Weight of the Stokes-vector Loss}
We test the weight of the loss function $w_i$ for the i-th Stokes vector in Equation~\eqref{eq:weight}.
Using the mean-squared-error (MSE) loss with the uniform weights of $w_i=1$ results in unstable training for the Stokes-vector elements with low values, e.g. the last element $[\mathbf{s}]_3$. 
Figure~\ref{fig:ablation}(a) shows that using our weighted loss of Equation~\eqref{eq:weight} allows for accurate reconstruction of the original Stokes vector by balancing the scale difference between the Stokes-vector elements.

\begin{figure*}[t]
	\centering
		\includegraphics[width=\linewidth]{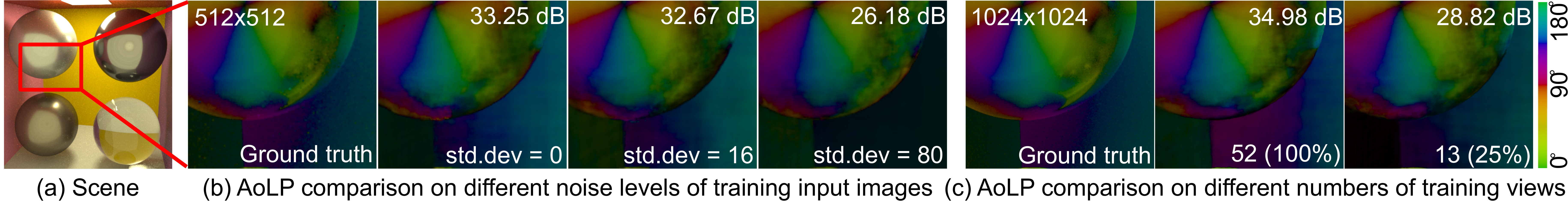}
		\caption{
{Input images with severe noise and fewer input views lead to performance degradation of NeSpoF.}
 }
\label{fig:results_synthetic_ablation}
\end{figure*}

\paragraph{Positional Encoding for Wavelength}
We test the impact of positional-encoding frequency $k$ of Equation~\eqref{eq:pos_enc} for the input wavelength $\lambda$.
Figure~\ref{fig:ablation}(b) shows that $k=1$ provides a high-fidelity plenoptic synthesis as visualized in the DoP and AoLP images. 

\paragraph{Measurement Noise}
The performance of NeSpoF is affected by measurement noise, as shown in rendering artifacts of low-SNR regions in real-world scenes such as the black background in Figure~\ref{fig:teaser}. 
We evaluate the impact of noise on NeSpoF with a synthetic scene with varying noise levels.
Figure~\ref{fig:results_synthetic_ablation}(a) shows that NeSpoF results in a minor PSNR drop of 0.58\,dB for the moderate noise level of Gaussian-Poisson noise with a standard deviation of 16. For the excessive noise level of standard deviation 80, the PSNR significantly drops by 7.07\,dB in PSNR.

\begin{figure}[t]
	\centering
		\includegraphics[width=\linewidth]{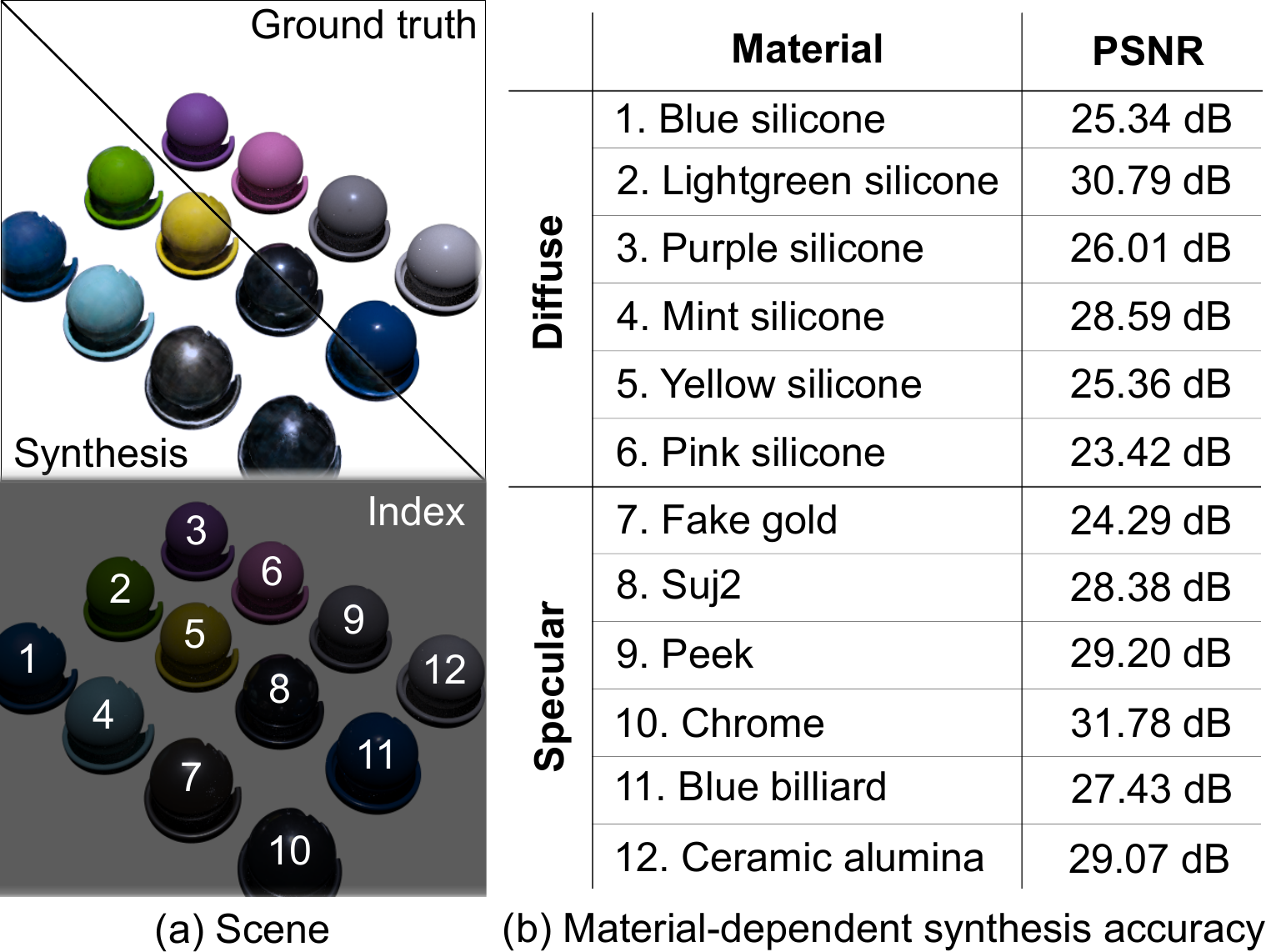}
		\caption{Impact of material appearance on NeSpoF. We used 12 materials from the multispectral pBRDF dataset~\cite{baek2020image}.
  }
		\label{fig:results_material_accuracy}
\end{figure}

\paragraph{Material Dependency}
The complexity of view-dependent spectro-polarimetric variation could vary with material types.
In Figure~\ref{fig:results_material_accuracy}, we evaluate NeSpoF on a synthetic scene consisting of multiple spheres with diverse materials.
The experiment shows varying PSNRs depending on material types. Developing a material-robust representation for modeling the complex interplay between light and spectro-polarimetric BRDFs of real-world materials~\cite{baek2020image} remains a topic for future work.

\paragraph{Image Resolution}
NeSpoF can handle high-resolution input images. 
Figure~\ref{fig:results_synthetic_ablation} shows an increase of PSNR by 1.73\,dB when using 1024$\times$1024 resolution compared to 512$\times$512 resolution.
However, we note that capturing such high-resolution images with a low level of measurement noise could be challenging. 

\paragraph{Number of Training Views}
The performance of NeSpoF depends on the number of training views. Figure~\ref{fig:results_synthetic_ablation}(b) shows that using 13 and 52 views for training results in the PSNRs of 28.82\,dB and 34.98\,dB, respectively. Improving NeSpoF for fewer views is an important problem, given the difficulty of capturing hyperspectral-polarimetric images at multiple views.

\section{Discussion}
\label{sec:discussion}
While NeSpoF aims to model additional wave properties of light, spectrum and polarization, NeSpoF still interprets light as rays, thus does not encompass phase and amplitude of light~\cite{goodman1996introduction}.
Also, NeSpoF demands hours of training time and tuning parameters that may benefit from recently-proposed neural representations. 
On the hardware front, our imaging system requires a lengthy capture time of 10 seconds for a single view. Also, the captured images contain considerable measurement noise due to the low transmission of the employed optical filters. We leave developing a single-shot light-efficient hyperspectral-polarimetric camera as a future direction.
Lastly, there is potential for further exploring the applications of NeSpoF in uncovering geometry, appearance, and material properties, similar to the progress on the applications of RGB radiance fields.

\section{Conclusion}
\label{sec:conclusion}
We propose NeSpoF, the first neural representation specifically designed for physically-valid hyperspectral Stokes-vector fields. Demonstrated across synthetic scenes and real-world scenes captured via a calibrated hyperspectral-polarimetric imager, NeSpoF reveals hidden spectro-polarimetric-positional information of the scene outside the confines of human vision. We anticipate that NeSpoF spurs further interest in high-dimensional imaging, modeling, and analysis.

\begin{acks}
This work was partly supported by Korea NRF (RS-2023-00211658, 2022R1A6A1A03052954, 2023R1A2C200494611), Samsung Research Funding Center (SRFCIT1801-52), Samsung Electronics, Korea MOTIE (NTIS1415187366, 20025752), and Korea IITP MSIT (No.2019-0-01906, Artificial Intelligence Graduate School Program-POSTECH).
\end{acks}


\bibliographystyle{ACM-Reference-Format}
\bibliography{references}


\end{document}